\documentclass[10pt,twocolumn,letterpaper]{article}

\usepackage{cvpr}
\usepackage{times}
\usepackage{epsfig}
\usepackage{graphicx}
\usepackage{amsmath}
\usepackage{amssymb}

\usepackage{iac_pkg}
\usepackage{tabulary}
\usepackage{multirow}
\usepackage{bm}
\usepackage{lipsum}
\usepackage{epigraph}
\usepackage{multirow}
\usepackage{gensymb}
\usepackage{tabularx, booktabs}
\newcolumntype{Y}{>{\arraybackslash}X}
\usepackage{pifont}
\usepackage[labelfont={bf,small},font=small]{caption}
\usepackage{subfig}

\usepackage{color}
\definecolor{citecolor}{RGB}{34,139,34}
\usepackage[pagebackref=true,breaklinks=true,letterpaper=true,citecolor=citecolor,colorlinks,bookmarks=false]{hyperref}
\usepackage[capitalise]{cleveref}

\cvprfinalcopy %

\def\cvprPaperID{} %

\begin{document}

\title{Towards Learning Structure via Consensus for Face Segmentation and Parsing}

\author{Iacopo Masi\qquad Joe Mathai \qquad Wael AbdAlmageed\\
USC Information Sciences Institute, Marina del Rey, CA, USA\\
{\tt\small \{masi,jmathai,wamageed\}@isi.edu}
}

\maketitle

\begin{abstract}
Face segmentation is the task of densely labeling pixels on the face according to their semantics. While current methods place an emphasis on developing sophisticated architectures, use conditional random fields for smoothness, or rather employ adversarial training, we follow an alternative path towards robust face segmentation and parsing. 
Occlusions, along with other parts of the face, have a proper structure that needs to be propagated in the model during training. Unlike state-of-the-art methods that treat face segmentation as an independent pixel prediction problem, we argue instead that it should hold highly correlated outputs within the same object pixels.
We thereby offer a novel learning mechanism to enforce structure in the prediction via consensus, guided by a robust loss function that forces pixel objects to be consistent with each other. 
Our face parser is trained by transferring knowledge from another model, yet it encourages spatial consistency while fitting the labels. Different than current practice, our method enjoys pixel-wise predictions, yet paves the way for fewer artifacts, less sparse masks, and spatially coherent outputs.
\vspace{-7pt}
\end{abstract}

\section{Introduction}\vspace{-5pt}

Face segmentation and parsing are invaluable tools since their output masks can enable next-generation face analysis tools, advanced face swapping~\cite{kemelmacher2016transfiguring,nirkin2018_faceswap,nirkin2019_fsgan}, more complex face editing applications~\cite{shu2017neural}, and face completion~\cite{liao2018face,li2017generative,mathai2019doesgenerative}.
Segmenting and parsing a face is strongly related to generic semantic segmentation~\cite{long2015fully,li2017fully,peng2017large,jegou2017one,lin2017refinenet,chen2018deeplab,Chen_2018_ECCV} since it involves the task of densely predicting conditioned class probabilities for each pixel in the input image according to pixel semantics. Although the two share the same methodology, face parsing is different than scene object segmentation since faces are already roughly scale and translation invariant, after a face detection step, and a plethora of methods has been developed towards solving the face parsing task~\cite{GLOC_CVPR13,liu2017face,liu2015multi,liu2011nonparametric,tsogkas2015deep,pan2018learning}.

\begin{figure}[t]
\centering
\includegraphics[width=\linewidth]{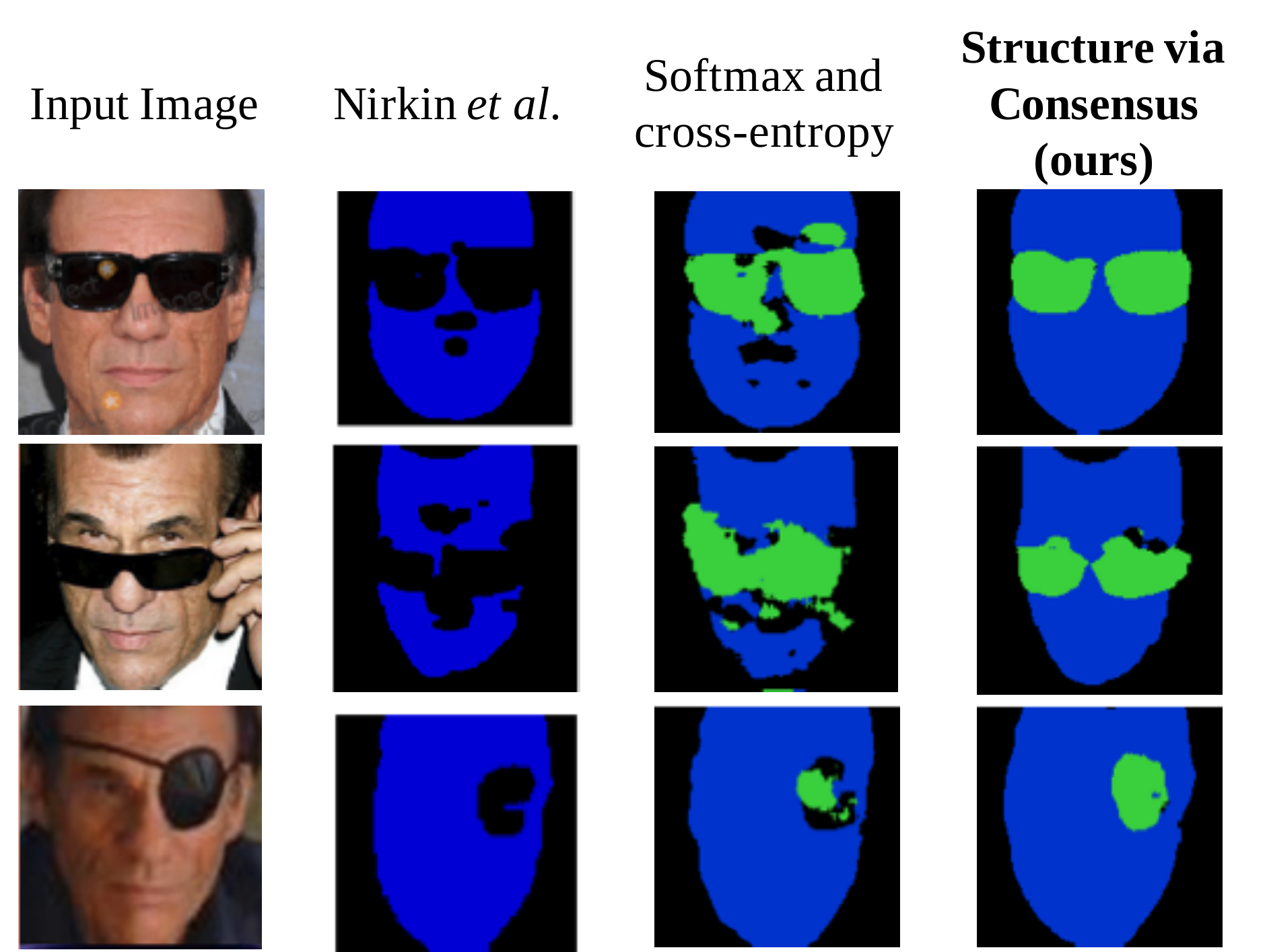}
\caption{
{\bf Structure via Consensus.} From left to right: the input image; result by~\cite{nirkin2018_faceswap}; our model with a pixel-wise loss; our proposed method. Previous methods predict only face (blue) vs. background (black), while ours yields separate predictions for occlusions (green) and background (black). Our loss enforces smoothness over objects covering the face via consensus constraint. %
\vspace{-10pt}
}
\label{fig:teaser}
\end{figure}

While state-of-the-art methods emphasize developing sophisticated architectures (e.g., two-stage networks with recurrent models~\cite{liu2017face}) or a complex face augmenter to simulate occlusions~\cite{nirkin2018_faceswap}, or rather employ adversarial training~\cite{pan2018learning}, we take an alternative path towards robust face segmentation and parsing.
Our method builds on an important observation related to the assumption of the \emph{independence} of pixel-wise predictions. Despite the significance of the aforementioned tasks, current methods overlook the regular structure present in nature and simply optimize for a cost that does not explicitly back-propagate any smoothness into the network parameters. This issue is particularly important for objects and faces, which  have a well-defined and continuous (non-sparse) structure.

\cref{fig:teaser} shows the advantage of the proposed method on a few samples drawn from the validation set tested on unseen subjects. While publicly available state-of-the-art models~\cite{nirkin2018_faceswap} perform face segmentation, they do so with very sparse and noncontinuous predictions and by modeling two classes only (face, non-face).  In contrast, by virtue of our method, we can separate occlusions from background, and more importantly, arrive at much more stable predictions that are hard to attain with a pixel-wise loss.

As also noted by~\cite{ke2018adaptive,hwang2018adversarial,liu2018coordconv}, training a network with pixel-wise softmax and cross-entropy for structured prediction makes the strong and too-simplistic assumption that pixel predictions are independent and identically distributed (i.i.d.). We take inspiration from the Gestalt laws~\cite{gestalt}---particularly the ones of proximity (close pixels shall be perceived as a group), closure (pixels shall be grouped into complete figures), good continuation (objects often minimize discontinuity)---and in response to the previous too-simplistic assumption, we make the following contributions which propose: (1) factorizing out occlusions by means of the difference between the complete face shape, attained through a strong prior robustly computed via 3D projections~\cite{chang17fpn,masi:icpr14}, and the output of a preexistent yet error-prone face segmentation network; (2) leveraging the connected components of the objects factorized before, using them as constraints to formulate a new loss function that still performs dense classification, yet enforces structure in the network by consensus learning; (3) finally showing that our approach is a generic tool for face parsing problems up to three classes, thereby reporting promising results in face parsing benchmarks~\cite{burgos2013robust,GLOC_CVPR13}. As an additional contribution, we have released our models and the related code\footnote{Available at \href{https://github.com/isi-vista/structure_via_consensus}{github.com/isi-vista/structure\_via\_consensus}}.

The remainder of this paper is organized as follows. \cref{sec:related} discusses related work, \cref{sec:method} explains our method, \cref{sec:expt} reports the experimental evaluation, and \cref{sec:conclusions} abstracts our findings along with future work.
\section{Related Work}\label{sec:related}
\minisection{Face segmentation} Recent work on face segmentation used a two-stream network~\cite{saito2016real} to predict a pixel-wise face segmentation mask. The system is fully supervised using pixel-wise segmentation masks obtained by preexisting data sets~\cite{LFWTech} or by additional semiautomatic manual efforts. Notably, \cite{saito2016real} is trained with pixel-wise softmax+cross-entropy, and in order to enforce regularization in the predicted mask, the method uses a conditional random field (CRF) as a post-processing step.
Importantly, CRFs have been already used in generic object segmentation and CNNs~\cite{zheng2015conditional,zhou2017face}. Adversarial learning has been used too for segmentation in~\cite{luc2016semantic}. Unlike all these methods, ours presents key differences in the way smoothness is propagated in the network.
Similar to~\cite{saito2016real}, Nirkin \etal ~\cite{nirkin2018_faceswap} trained a simple fully convolutional net (FCN~\cite{long2015fully}) for binary face segmentation using a semi-supervised tool to support manual segmentation of faces in videos; in our method we transfer knowledge from the weights of~\cite{nirkin2018_faceswap}, yet we demonstrate that by using our method we can learn from their mistakes and improve the model. Finally, Wang \etal \cite{wang2018face} exploited temporal constraints and recurrent models for face parsing and segmentation in video sequences and proposing a differentiable loss to maximize intersection over union (IoU)~\cite{rahman2016optimizing}. Other works extended the face segmentation problem to fine-grained face parsing~\cite{smith2013exemplar,jackson2016cnn,lin2019face}. %

\minisection{Semantic segmentation} Generic semantic segmentation has been an interesting topic in computer vision for a long time---starting with the seminal work using CRFs~\cite{boykov2001interactive,szummer2008learning} and graph cut~\cite{boykov1999fast,a22001fast}. CRFs impose consistency across pixels, assessing different affinity measures and solving the optimization through a message-passing algorithm~\cite{ross2011learning}. They have been successfully and widely used in face parsing applications also~\cite{GLOC_CVPR13,liu2015multi}. Recently, they began to be used as a post-processing step~\cite{saito2016real,liu2015multi,chen2018deeplab} with convolutional networks and later on expressed as recurrent neural networks~\cite{zheng2015conditional}. Super-pixels have also been employed to ease the segmentation process~\cite{fulkerson2009class,GLOC_CVPR13}, though  recently, the field was revolutionized with end-to-end training of FCNs,~\cite{long2015fully} optimized simply by extending a classification loss~\cite{AlexNet} to each pixel \emph{independently}. After~\cite{long2015fully}, there has been extensive progress in deep semantic segmentation--- mainly improving convolution to allow for wider receptive fields with its atrous (dilated) version~\cite{yu2015multi,Yu_2017_CVPR}, different spatial pooling mechanisms, or more sophisticated architectures~\cite{li2017fully,peng2017large,jegou2017one,lin2017refinenet,chen2018deeplab,Chen_2018_ECCV}.  

\minisection{Structure modeling} Modeling structure in computer vision dates back to perceptual organization~\cite{mohan1989using,sarkar1993perceptual,deutsch_ssvm15,deutsch_ijcai15,deutsch_singularity17,deutsch2019finding} and to the more general idea of describing objects with a few parts, advocating for  frugality~\cite{binford1971} in the shape description.
Lately, with modern deep-learning, in addition to  the aforementioned CRF formulation, all those concepts have faded away in the community---with some exceptions~\cite{tulsiani2017learning,ke2018adaptive}---and  instead  adversarial training~\cite{luc2016semantic,hung2018adversarial,ramirez2018exploiting,hwang2018adversarial} has been used to impose structure in the prediction forcing the output distribution to match the distribution of ground-truth annotations. %
Other attempts incorporate boundary cues in the training process~\cite{bertasius2016semantic,chen2016semantic} or pixel-wise affinity~\cite{bertasius2017convolutional}; others~\cite{jackson2016cnn} used a CNN cascade guided by landmark positions.
For an in-depth discussion on structured prediction, we refer to~\cite{nowozin2011structured}.
\section{Face Parsing with Consensus Learning}\label{sec:method}
Our objective is to robustly learn a nonlinear function $\phi$ parametrized by the weights of a convolutional neural network that maps pixel image intensities $\mbf{I} \in \mathbb{R}^{3\PLH H \PLH W}$ to a mask that represents per-pixel semantic label probabilities of the face $\mbf{y} \in \mathbb{R}^{K\PLH H \PLH W}$.
More formally, we aim to optimize $\phi(\mbf{I})$ so that it maps $\phi : \mathbb{R}^{3\PLH H \PLH W} \rightarrow \mathbb{R}^{K \PLH H \PLH W}$ where $K$ is the number of classes considered in our problem. Importantly, in the learning of $\phi(\cdot)$, while we minimize the expected cost across the training set, we need to enforce a mechanism that incorporates structure through smoothness. At test-time, like current practice, we obtain a final, hard-prediction as $\by^{\prime}\doteq \on{arg\,max}_k \phi(\mbf{I})$ and $\by^{\prime} \in \mathbb{R}^{1 \PLH H \PLH W}$.%

The following sections discuss how to obtain some external constraints for enforcing smoothness during the training, though later on we show that our method can be easily employed for the generic face parsing task. We do so by means of transferring knowledge from an existing network and using a strong prior given by 3D face projection to factorize out occluding blobs (\cref{sec:tk}).
Those blobs are then used to develop a novel loss function that instills structure via consensus learning (\cref{sec:consensus}).
 
\subsection{Face Segmentation Transfer} \label{sec:tk}

\minisection{Transfer data} Unlike~\cite{saito2016real} that took advantage of an existing yet small labeled set, or~\cite{nirkin2018_faceswap} that developed tools to assist the manual labeling, we use facial images from the CASIA WebFaces~\cite{yi2014learning}, VGG Faces~\cite{parkhi2015deep} and MS-Celeb-1M~\cite{guo2016ms} to harvest occlusions in-the-wild without any human effort. We argue that manually annotating them pixel-wise is a painstaking effort and practically infeasible. To pre-train our model, we used $539,960$ training images and $58,306$ validation images without overlapping subjects.
In the following sections we explain how we coped with the ambiguous and noisy synthesized pseudo-labels.

\minisection{Factorizing out occlusions} We express the occlusion appearing in a face image $\bI$ as the residual $\boldsymbol{\rho}$ obtained from the difference between the full face contour mask $\bff$ and the face segmentation mask $\mbf{f}^{occ}$ provided by~\cite{nirkin2018_faceswap}. More formally, given $\bI$ we further segment it as:
\begin{equation}
\boldsymbol{\rho} = \hinge{\bff - \mbf{f}^{occ}},\quad \on{where}\hinge{\cdot} = \on{max}\big(0,\cdot \big).%
\label{eq:res}
\end{equation}
\cref{eq:res} serves to factorize the occlusions out from the background. The mask $\bff$ is expressed as the convex hull of the full face shape predicted by projecting onto the image a generic face shape via 3D perspective projection~\cite{masi:icpr14} computed using the robust method mentioned in~\cite{chang17fpn}. Note that since we are interested in the facial outer contour, \cite{chang17fpn} fits our needs since it favors robustness to precision---which is especially useful in the presence of occlusions. This is easily implemented by obtaining the predicted pose $\on{K}[\on{R}|\on{t}]$ and projecting 64 vertices onto the image corresponding to the 3D contour of the face (jawlines plus forehead). Then, $\bff$ is efficiently computed finding the simplex of the convex-hull and probing to find if a matrix index $s=(i,j)$ of $\bff$ is outside the hull, where $s$ runs over all the pixels in an image.
By construction, the residual takes values in $\{-1,0,+1\}$ and is then truncated to $\{0,1\}$ as stated in \cref{eq:res} to remove possibly ambiguous labels.
\begin{figure}[t]
\centering
\includegraphics[trim={0 15pt 0 20pt},clip,width=\linewidth]{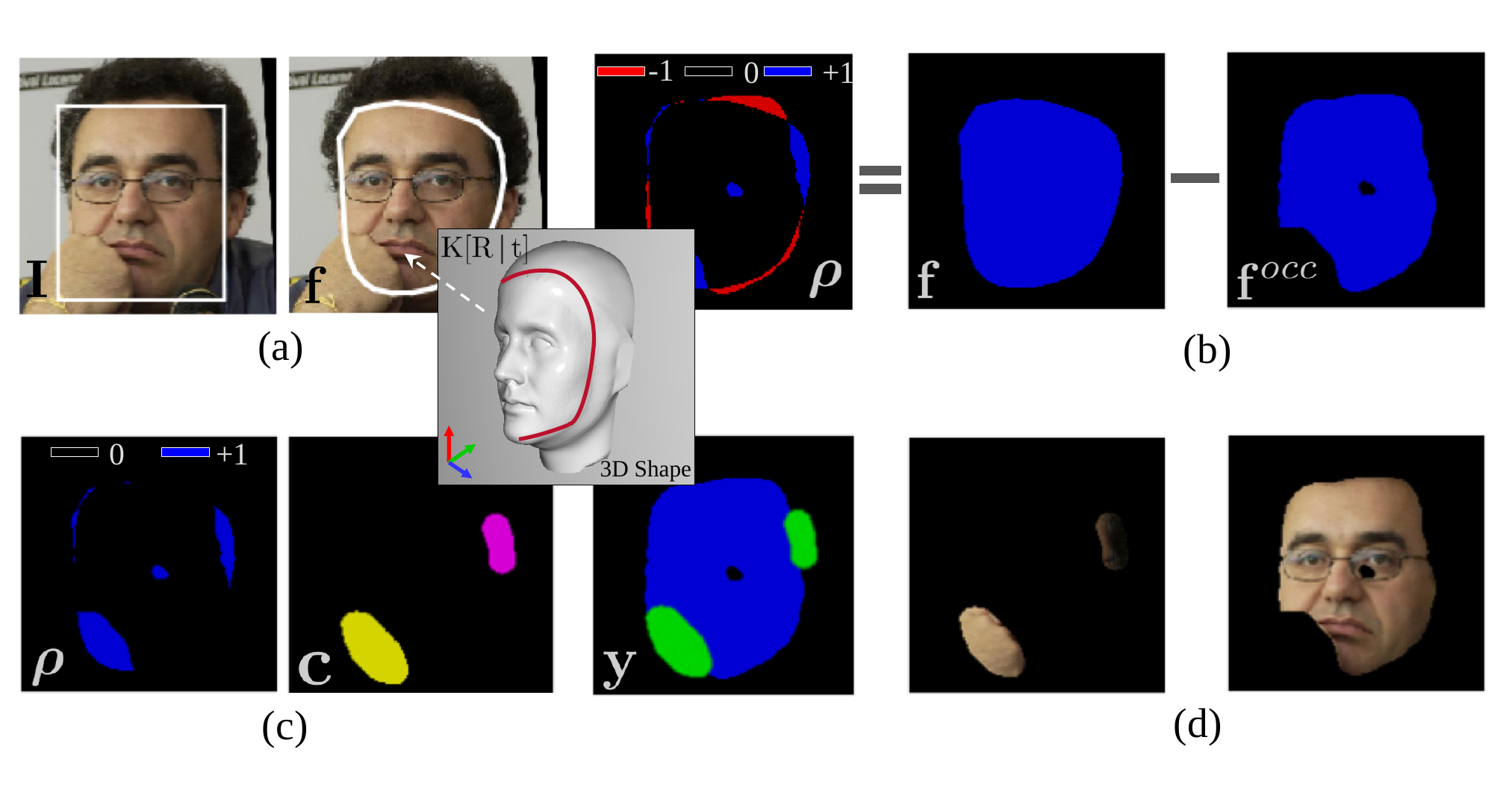}
\vspace{-20pt}
\caption{
{\bf Factorizing out occlusions.} (a) We use face detection and a strong prior provided by the projection of a 3D facial shape to get the full face mask; (b) The initial $\boldsymbol{\rho}$ residual is expressed as the difference between the full and the segmentation mask; (c) $\boldsymbol{\rho}$ is refined and its connected components $\bc$ estimated; (d) Label $\by$ is obtained from $\bc$ to decouple the occlusions from the face.
\vspace{-10pt}
}
\label{fig:preproc}
\end{figure}
The residual then undergoes a series of morphological operations to amplify the occlusions, since, for example, in face completion applications~\cite{liao2018face,li2017generative} over-segmentation of occlusions is preferable over under-segmentation. The final $\boldsymbol{\rho}$ is obtained by applying an erode operator twice with rectangular kernels of size $25\PLH7$ and a dilation operation with elliptical kernel of size $45\PLH45$. The values are chosen to be conservative with respect to the occlusions: in case the teacher network undersegments occlusions, the rationale was to amplify the occlusions over the face regions.
Finally, the connected components are estimated from the residual to identify main blobs or objects on the face. By merging the output of the face segmentation network $\mbf{f}^{occ}$ and the labels provided by the connected components, the method yields a pseudo-ground-truth mask $\bc(\bI) \in \mathbb{R}^{1\PLH H\PLH W}$, where $\bc$ takes values in $\big\{0,\ldots,N_C\big\}$. Note that, $N_C$ is not constant since the number of blobs---i.e., connected components---varies across images; yet by construction, we have that the pixel-wise semantic labels $\by$ are defined as:

\begin{equation}\label{eq:residual_value}
\forall s,\quad \by(s) = 
\begin{cases} 
s \sim \on{background},       & \text{if}~\bc(s) = 0, \\ 
s \sim \on{face},          & \text{if}~\bc(s) = 1, \\  
s \sim \on{occlusion},        & \text{if}~\bc(s) \geq 2.   
\end{cases}
\end{equation}
The entire process is summarized in \cref{fig:preproc}.

\subsection{Enforcing Structure via Consensus Learning} \label{sec:consensus}

\minisection{Network structure} We employ a simple network based on a fully convolutional encoder-decoder~\cite{ronneberger2015u} taking as input $128\PLH128$ RGB images. We note here that the goal is not having a state-of-the-art architecture but to prove the effect of our regularization on the smoothness of the masks. 
The network uses recurrent applications of a basic building block of Conv--Elu--BatchNorm~\cite{clevert2015fast}. The model has two encoding branches: a first encoding branch increases the depth while decreasing spatial dimension up to $256\PLH32\PLH32$. %
The second sub-encoder refines the feature maps of the first encoder focusing the attention on a wider part of the input face, using two blocks with dilated convolutions~\cite{yu2015multi}. The feature maps of the two encoders are concatenated together. The decoder maps back to the input spatial dimension using efficient sub-pixel convolution~\cite{shi2016real} with upscaling
ratio of two to upscale the feature maps.
Importantly, a final pixel in the classification layer has a receptive field in the input image of $121$ pixels, hence it almost covers the entire face\footnote{For additional details on the network architecture please check the supplementary material.}. 

\minisection{A critique of pixel-wise loss functions}
The general recipe for semantic segmentation boils down to transforming an image $\mbf{I}$ using a network $\phi$ that generates a $K\PLH H\PLH W$ tensor of probabilities $\mbf{p}$ to maximize the conditioned probability given the ground-truth mask $\mathbf{y}$ with size $1\PLH H \PLH W$. The network output is expressed as a set of multinoulli\footnote{Generalization of Bernoulli distribution with K categories, also known as categorical distribution.} distributions, where each pixel prediction $s\sim \on{Cat}\big(K,\ps\big)$.
The fitting to the mask labels is implemented with pixel-wise softmax plus cross-entropy, finally averaged over the final tensor. This introduces a strong assumption: all the final generated pixels in the mask behave as independent and identically distributed (i.i.d.) random variables, which violates the regular structures implicitly present in nature~\cite{hwang2018adversarial}.
Defining a pixel location as $s$, the expected loss across all pixel's image $\mathbb{E}[\ell(\cdot)]$ is eventually:
\begin{multline}\label{eq:softmax} 
\frac{1}{HW}\sum_{s=1}^{HW}  \ell(s) = \frac{1}{HW}\sum_{s=1}^{HW} \mathcal{H}\Big(\ps, \mbf{y}(s)\Big) = \\
= - \frac{1}{HW}\sum_{s=1}^{HW} \mbf{y}(s)\log\big(\ps\big)~,
\end{multline}
where $\mathcal{H}$ indicates the cross-entropy between the predicted softmax probability $\ps$ and $\mbf{y}(s)$ is one-hot encoding of the class membership at a pixel $s$. More analytically:
\begin{equation}
\mathbb{E}[\ell(\cdot)] = -\frac{1}{HW}\sum_{s \in \by} \log\Bigg(  \frac{e^{ \mbf{W}_{(k^{\star},s)}\mbf{x}_{(k^{\star},s)}^T}  }{\sum_{k=1}^K e^{\mbf{W}_{(k,s)}\mbf{x}_{(k,s)}^T }}
\Bigg),
\label{eq:softmax_an}    
\end{equation}
where $k$ runs over the classes, $k^{\star}$ selects the ground-truth class index based on $\mbf{y}$, and $s$ runs on all the pixels. $\mbf{W}$ represents the final classification convolutional layer mapping to the label space and $\mbf{x}$ the activation before $\mbf{W}$.

\cref{eq:softmax_an} assumes that the prediction at a given pixel is not regularized by the structure present in the input, and hence it suggests improvement by incorporating smoothness constraints. Although each pixel prediction in $\mathbf{x}$ has some knowledge of the neighbour pixels in the input image, given the recurrent application of convolutions, this is not enough to avoid predicting pixels independently, even in the dilated case~\cite{yu2015multi,Yu_2017_CVPR} allowing for large receptive fields as in our model. %
Despite the recent progress in semantic segmentation~\cite{Chen_2018_ECCV}, the aforementioned issue is not yet fully addressed in the face domain. \cref{eq:softmax_an} is also often used in applications such as face segmentation, face parsing or occlusion detection, and in many cases where the network has to densely label pixels. The problem of returning sparse predictions is especially important on faces, since these exhibit a very regular structure. The same is true for occlusion covering the face: obstructing objects covering the face are rarely composed of sparse tiny parts, yet rather show up with continuous shapes.

\minisection{Preliminaries} The above problem calls for a solution regarding the independent assumption of the predictions in \cref{eq:softmax_an}. Unlike~\cite{saito2016real,nirkin2018_faceswap} that couple the background and occlusion classes together, we define face segmentation as a three-class problem ($K=3$) aiming to classify background $\mbf{B}$, face $\mbf{F}$ and occlusion $\mbf{O}$. Additionally, following \cref{sec:tk}, we allow for occlusions to be modeled as a variable set of blobs over the face $\mathcal{O}=\{\mbf{O}_1,\ldots,\mbf{O}_n\}$.
In spite of this,  \cref{eq:softmax} can be rewritten as:
\begin{equation}
\frac{1}{|\mbf{F}|}\sum_{s\in \mbf{F}} \ell(s) + \frac{1}{|\mbf{B}|}\sum_{s\in \mbf{B}} \ell(s) + \frac{1}{|\mathcal{O}|}\sum_{\mbf{O}\in\mathcal{O}} \Bigg( \frac{1}{|\mbf{O}|}\sum_{s\in \mbf{O}} \ell(s) \Bigg),
\label{eq:softmax_blobwise}    
\end{equation}
where $\ell(s)$ corresponds to the softmax plus cross-entropy loss at a pixel $s$, that runs over all the pixels in each blob, and $|\cdot|$ counts the pixels of a blob.
\cref{eq:softmax_blobwise} is identical to \cref{eq:softmax}, with the only difference being that the spatial frequency of each component is marginalized out, or in other terms, having the same weights for all blobs irrespective of their size.  Next, we explain how to enforce smoothness in our training process.%

\begin{figure}[t]
\centering
\includegraphics[width=\linewidth]{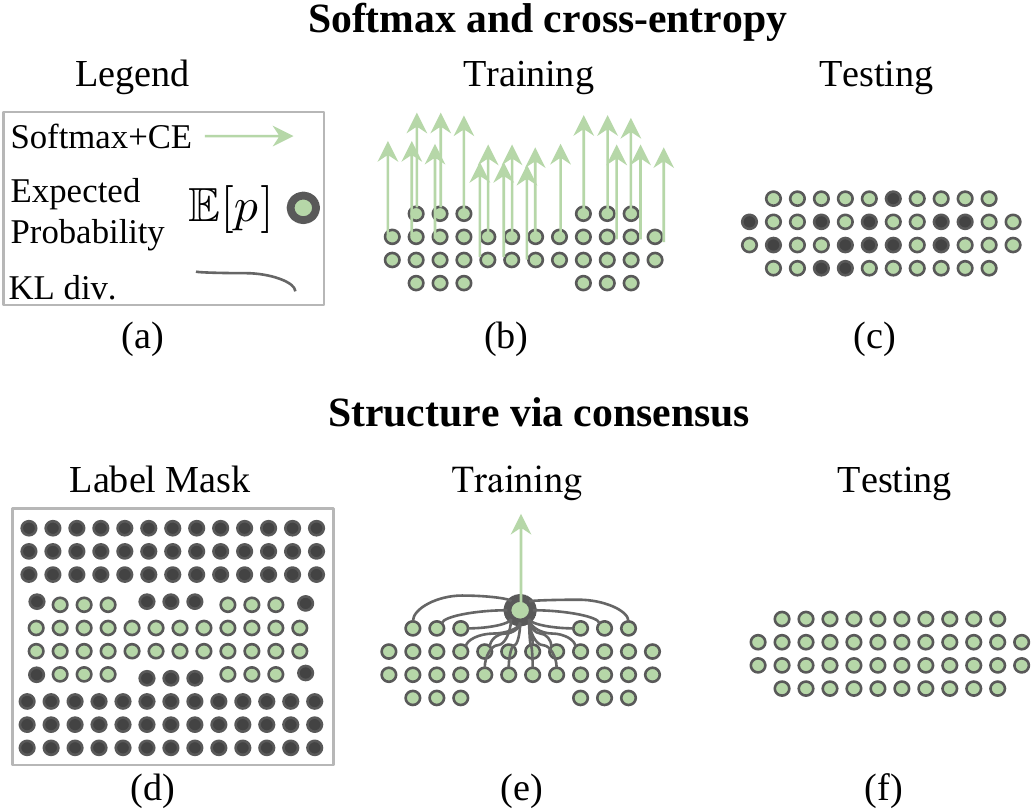}
\caption{
{\bf Intuition behind our loss.} (a) legend; (b) regular training proceeds pixel-wise, \emph{independently} and enforces densely each pixel to fit the label (no notion of smoothness of the object); (c) this leads to sparse prediction at test time for unseen objects; (d) pixel-wise labels for an image; (e) we force the expected prediction in a blob $\mathbb{E}[\bp]$ to the label, yet ensure no deviation of each pixel from the average; (f) the network is better regularized for segmenting \emph{with less sparse predictions}.
\vspace{-15pt}
}
\label{fig:loss}
\end{figure}

\minisection{Enforcing structure in each blob} 
The core idea behind our method is shown in \cref{fig:loss}. We define the expected probability $\mathbb{E}[{\bp}]$ on a blob $\mbf{c}$ as:
\begin{equation}\label{eq:expect_prob_blob}
\widehat{\bp}_{\mbf{c}} \doteq  \frac{1}{|\mbf{c}|}\sum_{s\in \mbf{c}} \bp(s) =\frac{1}{|\mbf{c}|}\sum_{s\in \mbf{c}}\Bigg(  \frac{e^{ \mbf{W}_{(k^{\star},s)}\mbf{x}_{(k^{\star},s)}^T}  }{\sum_{k=1}^K e^{\mbf{W}_{(k,s)}\mbf{x}_{(k,s)}^T }}\Bigg),
\end{equation}
that corresponds to the average conditioned probability over all the pixels of the blob $\mbf{c}$. Note that the values in \cref{eq:expect_prob_blob} remain positive and the mass of $\widehat{\bp}_{\mbf{c}}$ sums up to one.
Then, we can augment \cref{eq:softmax} in the following way: given a blob on the mask we can define the loss on the blob as
\begin{equation}
\underbrace{\alpha\cdot D_{KL}\big(\mbf{y}~||~\widehat{\bp}_{\mbf{c}}\big)}_{\text{1st order; matches the class label}} + \underbrace{\beta\cdot\frac{1}{|\mbf{c}|}\sum_{s\in \mbf{c}} D_{KL}\big(\widehat{\bp}_{\mbf{c}}~||~\ps\big)}_{\text{2nd order; ensures no deviation}},
\label{eq:avg_dev_1}
\end{equation}
where $\alpha$, $\beta$ are two constant parameters controlling the trade-off between matching the labels and ensuring consensus and $D_{KL}$ denotes the Kullback-Leibler divergence. %
Following the notation of ~\cref{sec:tk}, and putting this all together, indicating all the the blobs (background $\mbf{B}$, face $\mbf{F}$, occlusions $\mathcal{O}$) as $\bc$, our method finally optimizes:
\begin{multline}
\frac{1}{|N_C|}\sum_{\mbf{c}\in \bc} \Bigg[  \alpha\,D_{KL}\big(\mbf{y}~||~\widehat{\bp}_{\mbf{c}}\big) +\frac{\beta}{|\mbf{c}|}\sum_{s\in \mbf{c}} D_{KL}\big(\widehat{\bp}_{\mbf{c}}~||~\ps\big) \Bigg].
\label{eq:avg_dev_2}
\end{multline}
Note that although here we apply our formulation specifically to face segmentation/occlusion detection, if the method is provided with a set of blobs, then it can be applied more broadly. \cref{sec:expt} shows how to easily obtain blobs from the available labels in benchmarks for a generalization to face parsing with a small number of classes.

\subsection{Interpretations} \label{sec:inter}
\cref{eq:avg_dev_2} can be interpreted as follows: given a blob on the mask $\mbf{c}$, we enforce that the average of the 
predictions over the blob has to match the class label $D_{KL}\big(\mbf{y}~||~\widehat{\bp}_{\mbf{c}}\big)$---as a sort of first-order momentum---plus a second term ensures that all pixel-wise probabilities inside the blob are close to its average, i.e., $\sum_{s\in \mbf{c}} D_{KL}\big(\widehat{\bp}_{\mbf{c}}~||~\ps\big)$. We treat each blob as a whole using the first term and we enforce smoothness using the regularization of the second term: unlike the baseline, our loss connects all the pixel predictions in a blob with the average prediction, defining implicit inter-dependencies as a sort of regularizer.
As a cross-topic parallelism, it may be useful to the reader to know that a similar smoothness regularization has been proposed recently to induce smoothness to cope with adversarial attacks~\cite{kannan2018adversarial}.

\minisection{Implementation} In the first term, what is actually implemented as $D_{KL}$ is the negative log-likelihood of the ground-truth probability from $\widehat{\bp}_{\mbf{c}}$. This can still be viewed as KL div. since this latter reduces to cross-entropy given that $D_{KL}\big(\mbf{y}~||~\widehat{\bp}_{\mbf{c}}\big) \doteq \mathcal{H}\big(\mbf{y},\widehat{\bp}_{\mbf{c}}\big) - \mathcal{H}\big(\mbf{y}\big)$, and, $\mbf{y}$, the target distribution, is a one-hot encoding, thus with entropy equal to zero. Hence, KL div. is equal to cross-entropy in this case. The second term in \cref{eq:avg_dev_2} is simply implemented as KL div. between two discrete distributions.
In this sense, \cref{eq:avg_dev_2} keeps an elegant consistency across its two terms, without requiring the system for external CRF post-processing or additional parameters to perform adversarial training.

\minisection{Interpretation as a generalization of \cref{eq:softmax_an}} Additionally, the proposed formulation can be seen as a generalization of \cref{eq:softmax_an}. A pixel-wise loss coincides with a boundary case of our loss when all the blobs collapse down to each pixel. In this case, each pixel matches the class label---first term in~\cref{eq:avg_dev_2}---and the second term collapses to zero, since, by definition, a pixel is consistent to itself.

\minisection{Connection to CRFs} Our formulation shares some similarities with seminal CRFs~\cite{boykov2001interactive,szummer2008learning,liu2015multi} and graph cut~\cite{boykov1999fast,a22001fast} for semantic segmentation. At first sight the two terms in \cref{eq:avg_dev_1} are reminiscent of minimizing the energy of a function $\phi$ as $E_{data}(\phi)+E_{smooth}(\phi)$, as proposed in~\cite{a22001fast}. Though the CRF has already been used in conjunction with a ConvNet (e.g.,~\cite{zheng2015conditional,chen2018deeplab}), we do share the core philosophy with novel traits; unlike~\cite{a22001fast}, our ``unary potential'' is not defined on single pixels but on the expected probability over the shape, and our ``pair-wise potential'' is not defined on pairs of \emph{adjacent pixels}~\cite{a22001fast} or \emph{fully connected}~\cite{chen2018deeplab}, yet is constrained by components with characteristic shapes. We note here that in our case $\phi$ is parameterized by the filters of a ConvNet. Finally, we acknowledge that CRFs captures long range interactions via a fully-connected graphical structure, in contrast, the proposed loss only captures constraints within neighborhoods; though, the ``neighborhood'' in our case can be small or large depending on the label masks or connected components mined in \cref{sec:tk}. In light of this, our formulation still exhibits innovative traits.%

\begin{figure*}[tb]
\centering
\subfloat[Training progress]{
    \includegraphics[trim={0 0 0 0},clip,width=0.25\linewidth]{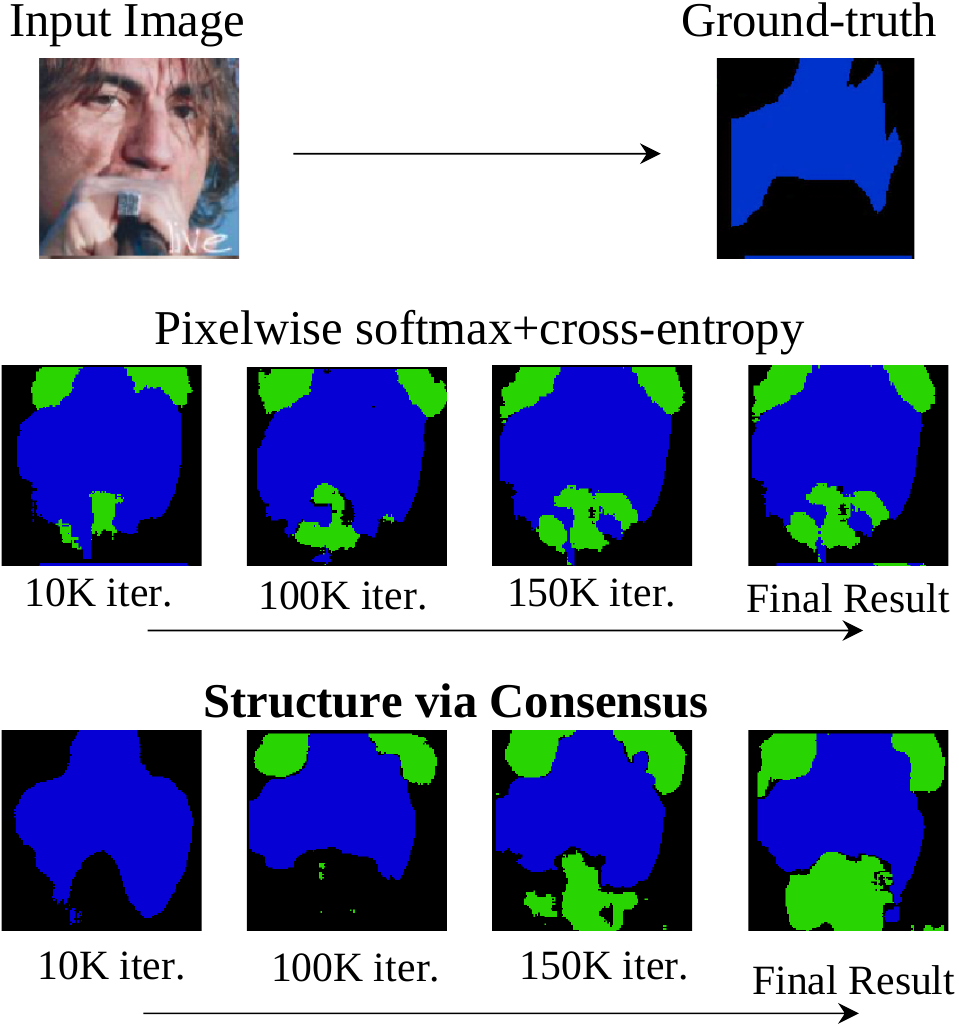}
    \label{fig:valid_loss:c}
}\qquad
\subfloat[$\vert\on{cc}_{\phi}-\on{cc}_{gt}\vert$]{
    \includegraphics[trim={0 0 0 0},clip,width=0.265\linewidth]{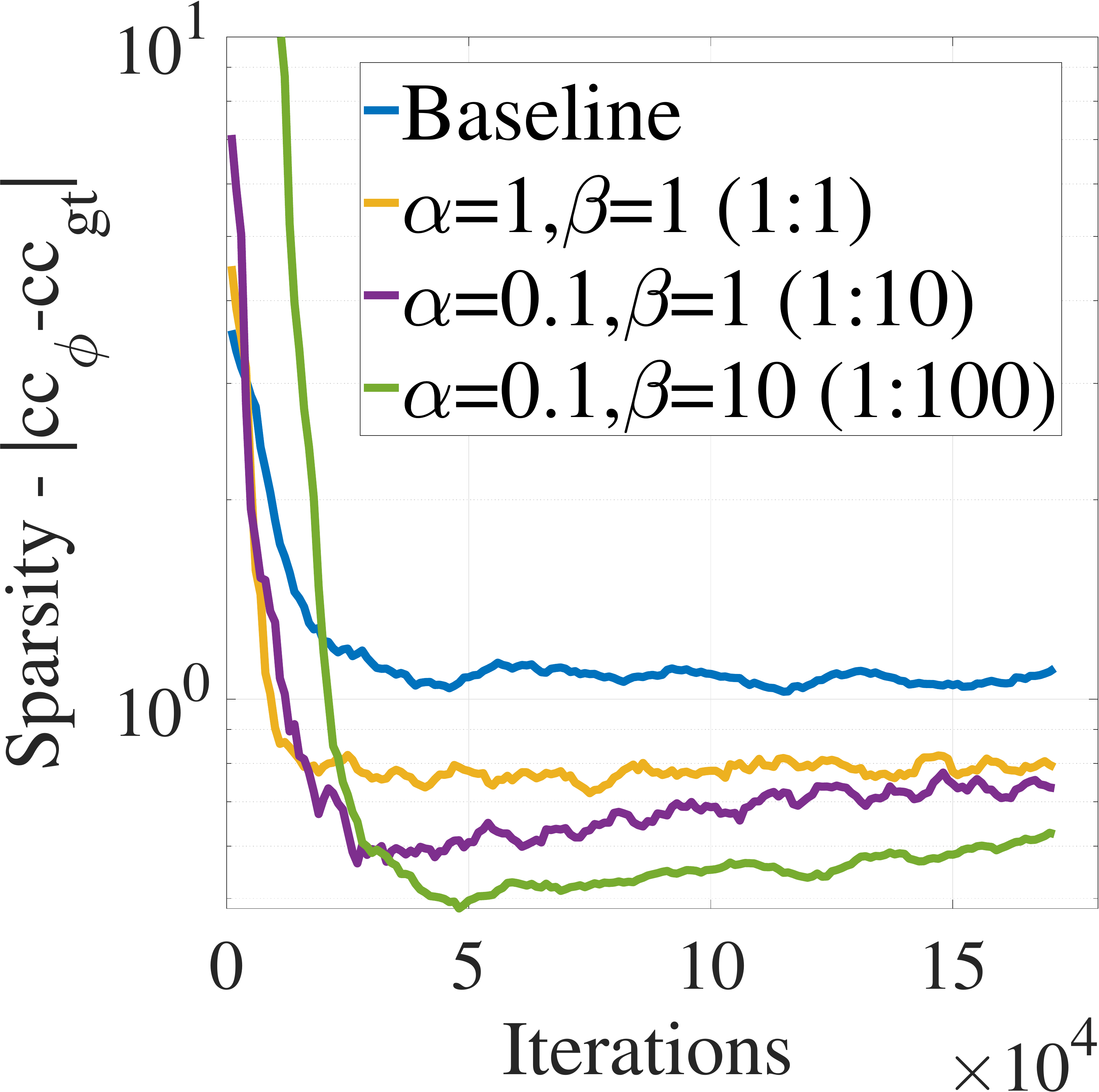}
    \label{fig:valid_loss:d}
}\qquad
\subfloat[Increasing $\beta$ reduces sparsity]{
\includegraphics[trim={0 0 0 0},clip,width=.35\linewidth]{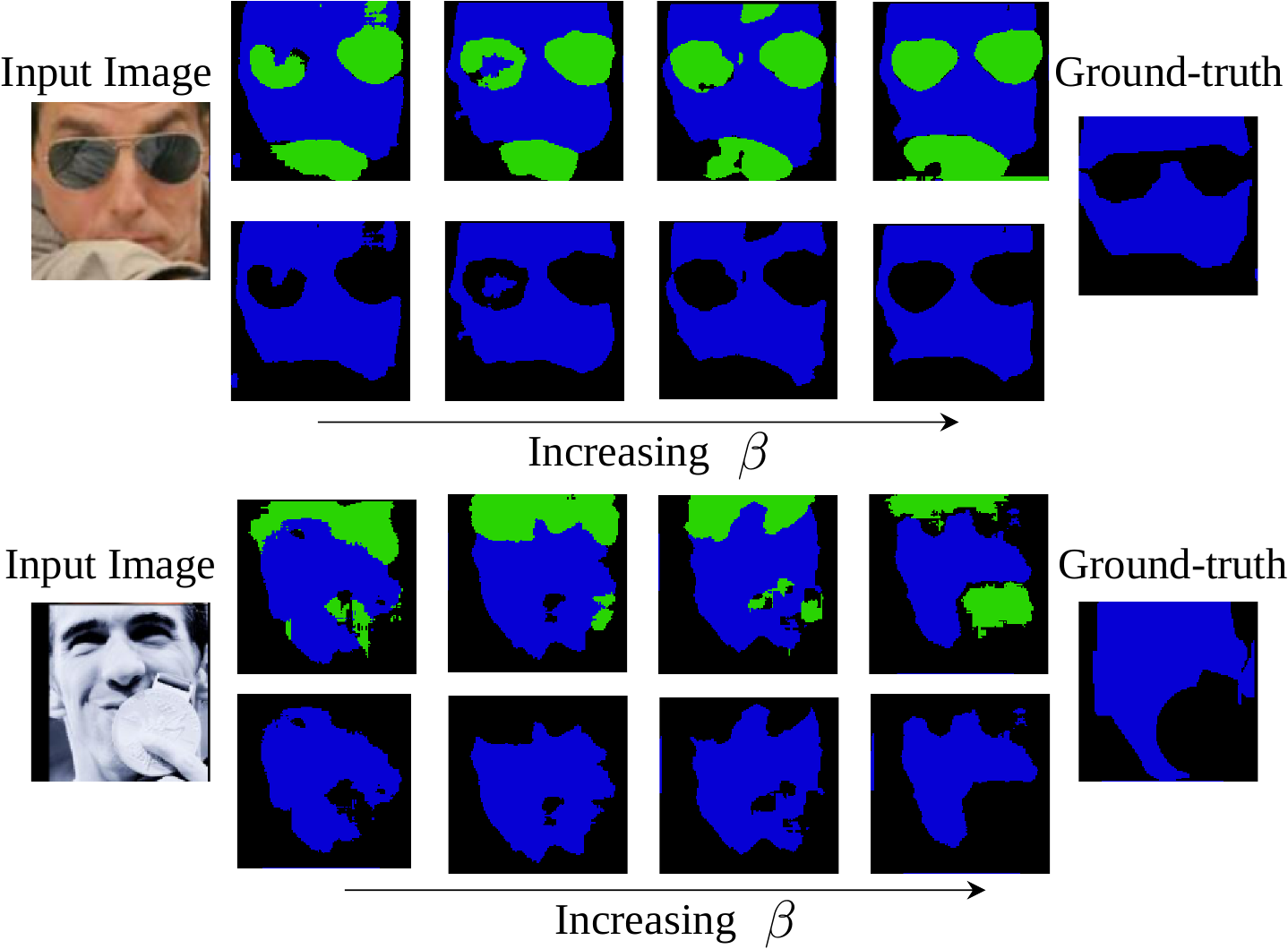}
\label{fig:qual_cofw_beta}
}
\vspace{-5pt}
\caption{
{\bf More regular, smooth structure learned}; (a) As the training progresses, our method learns more regular, smooth structure which yields a more regular mask when compared to the pixel-wise baseline (sample from the COFW test set); (b) less sparsity is confirmed by visualization of the error in the number of connected components between the predicted $\on{cc}_{\phi}$ and annotated mask $\on{cc}_{gt}$.  A higher weight on $\beta$ greatly decreases the sparsity of the masks (c) this effect is confirmed when inspecting qualitative samples from the COFW test set.
\vspace{-10pt}
}
 \label{fig:valid_loss}
\end{figure*}

\section{Experimental Evaluation}\label{sec:expt}
We report results of ablation study or experiments that motivated our choices, along with state-of-the-art evaluations on benchmarks for face segmentation, occlusion detection and face parsing. Our approach surpasses previous methods by wide margins on the COFW (Caltech Occluded Faces in the Wild)~\cite{burgos2013robust} and shows comparable results on the Part Labels set~\cite{GLOC_CVPR13} despite using a lightweight model.

\subsection{Implementation Details} \label{sec:impl}
\minisection{Face preprocessing} Following~\cite{liu2017face}, we used a minimalist face preprocessing, simply applying a face detector~\cite{yang2016multi} and using the adjusted square box to crop and resize each face and its corresponding label to $128 \PLH 128$ pixels. On \partlabels, faces are aligned thus we just resize them to $128$p.

\minisection{Training} To pre-train the network we use the Adam optimizer~\cite{kingma2014adam}, starting from a learning rate of $\on{1e-3}$ and finishing with $\on{1e-5}$. Pseudo-labels are provided following \cref{sec:tk}. A scheduler checks the pixel-wise average recall across classes on the validation and decreases the learning by $\on{1e-1}$ when the above metric plateaus. %
All the models are trained with a batch size of $128$. When fine-tuning on COFW, we apply our face segmentation transfer (\cref{sec:tk}) to identify the main blobs without applying the morphological operations to use the fine-grain human annotated masks. In other tests, we simply treat the separate mask classes as the blobs. On COFW we used a flat learning rate of $\on{1e-5}$, while on \partlabels~$\on{1e-4}$. All the models are fine-tuned until convergence reaches saturation on the training set. Important parameters in \cref{eq:avg_dev_1} are $\alpha$, $\beta$ that are set as $\{10$:$5\}$ in all our experiments, as we found these values to be a good trade-off between enforcing smoothness and fitting the labels to guarantee high accuracy.

\subsection{Supporting Experiment}

\minisection{More regular, less scattered structure} \cref{fig:valid_loss:c} shows qualitatively the difference in the prediction between the baseline and learning with \emph{structure via consensus} on a COFW~\cite{burgos2013robust} test sample when performing transfer learning with our loss. The sample is chosen for its difficulty in the face segmentation task (the occlusion appears fragmented---although it is not---and is of similar color to the face, in spite of the fact that it is made by two continuous objects (e.g., hands and microphone). As the training progresses, our method offers more continuous segmentation masks that, in turn, become a better face segmentation, without sparse holes. Our claim is supported by \cref{fig:valid_loss:d}, showing the average absolute error $\frac{1}{N}\sum_i^N \vert\on{cc}_{\phi}-\on{cc}_{gt}\vert$ between the number of connected components in the ground-truth mask ($cc_{gt}$)  and the components dynamically computed on our prediction ($cc_{\phi}$) at every iteration. The error is averaged across all the testing samples and provides a valuable understanding of the sparsity of the prediction and confirms that increasing our smooth term $\beta$ in \cref{eq:avg_dev_2} induces a significant less sparse output. \cref{fig:valid_loss:d} shows the trend of the sparsity error measure as the training evolves for different $\beta$ values. Additional qualitative samples in \cref{fig:qual_cofw_beta} further support our hypothesis.

\begin{figure}[tbh]
    \centering
     \small{\hspace{-12pt}Input, Label Mask\hspace*{4pt}
     Nirkin \etal~\cite{nirkin2018_faceswap} \hspace*{11pt}
     Baseline \hspace*{15pt}
     \tbf{Ours}\\}
    \includegraphics[width=0.85\linewidth]{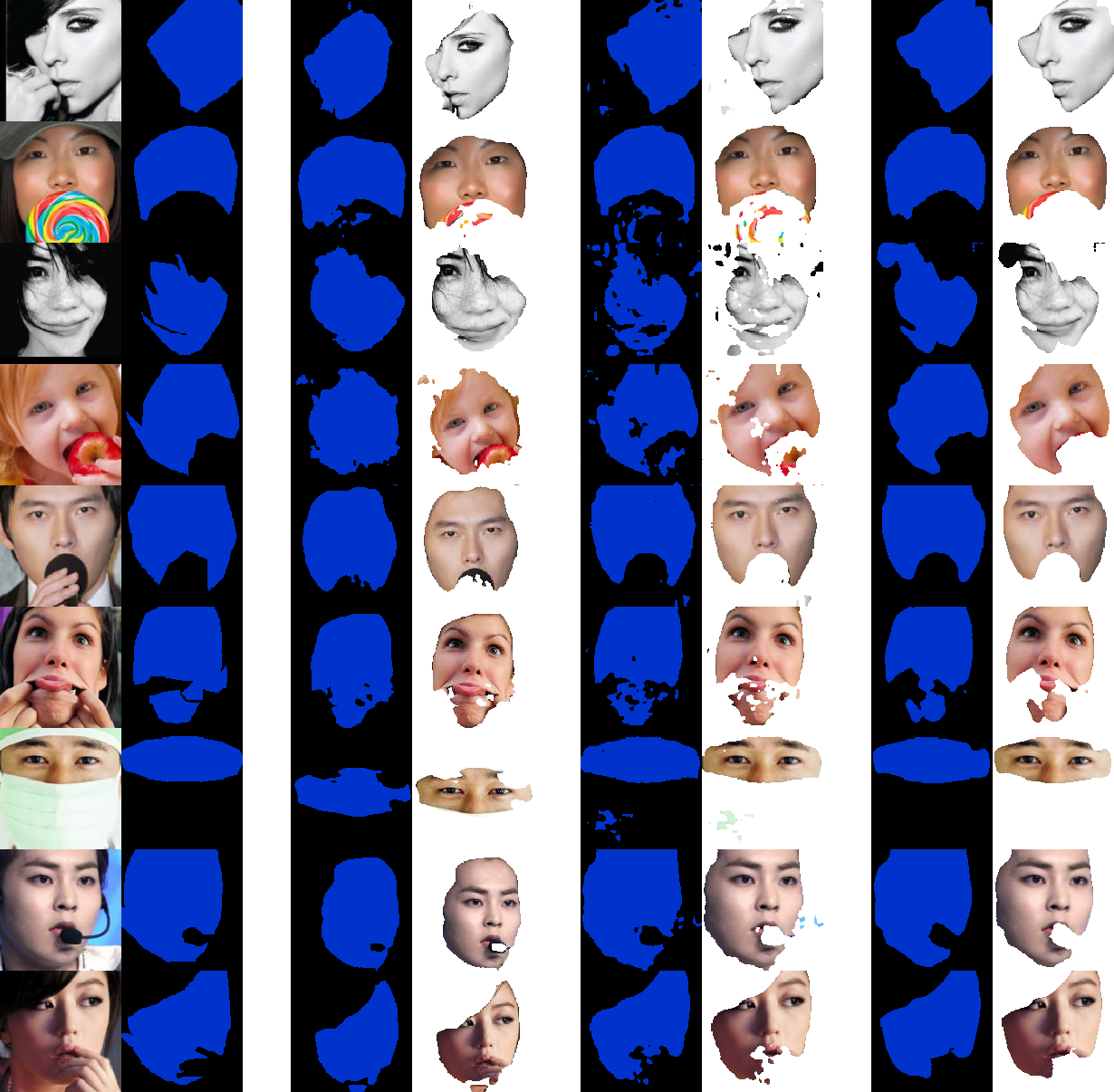}
    \caption{{\bf Qualitative samples from the COFW set.} Input image and its ground-truth mask; results by Nirkin \etal~\cite{nirkin2018_faceswap}; baseline with pixel-wise loss; our result. The faces are masked to remove occlusions according to each method. Additional results in the supplementary material in \cref{fig:cofw1}. 
    \vspace{-10pt}}
    \label{fig:cofw_qual}
\end{figure}

\subsection{Caltech Occluded Faces in the Wild} \label{sec:cofw}

\begin{table}[t]
\centering
\resizebox{1.0\linewidth}{!}{
\begin{tabular}{l c@{~}c@{~}c@{~}c@{~}c@{~}c}
\toprule
Method & IOU$_\text{\emph{face}}$ & acc. & rec$_\text{\emph{face}}$ & rec$_\text{\emph{all}}$ & spars. &  fps  \\
\cmidrule(r){1-1} \cmidrule(l){2-7}
Struct. Forest~\cite{jia2014structured} & --- & 83.9 & --- & 88.6 & --- & --- \\
RPP~\cite{Yang_tip_land} & 72.4 & --- & --- & --- & --- & 0.03 \\
SAPM~\cite{ghiasi2015using} & 83.5  & 88.6 &  87.1 & --- & --- &   --- \\
Liu \etal~\cite{liu2015multi} & 72.9 & 79.8 & 89.9 &  77.9 & --- &   0.29\\
Saito \etal~\cite{saito2016real} \emph{\small{+GraphCut}} & 83.9  &  88.7 &  92.7 & --- & --- &  43.2 \\
Nirkin \etal~\cite{nirkin2018_faceswap} & 81.6 & 87.4 & 93.3 & --- & --- &  48.6\\
Nirkin \etal~\cite{nirkin2018_faceswap} \emph{\small{+Occ. Aug}}.  & 83.7  & 88.8 & {\bf 94.1}  & 87.4 & --- & 48.6\\
\cmidrule(r){1-7}
Softmax+CE \emph{\small{+Scratch}}          & 76.8 & 83.7 & 86.9 & 82.6 & 3.5 & 300\\
\cmidrule(r){1-1}
Softmax+CE \emph{\small{+Transf.}}             & 84.5 & 89.4 & 93.3  &  88.1 & {1.0} &  {300}\\
\cmidrule(r){1-1}
Softmax+CE \emph{\small{+Transf.+f.t.}}               & 84.1 & 89.4 & 90.3 & 89.1   & 3.8  &   {300}\\ %
Struct. via con. \emph{\small{+Transf.+f.t.}}  & {85.7} & {90.4} & {92.5} & {89.7} & 1.6 & {300}\\
\cmidrule(r){1-1}
Struct. via con. \emph{\small{+Transf.+f.t.+reg.}}  & \tbf{87.0} & \tbf{91.3} & {92.4} & \tbf{90.9} & \tbf{0.8} & \tbf{300}\\
\bottomrule
\end{tabular}
}
\caption{{\bf COFW set.} Occlusion segmentation results.\vspace{-5mm}}%
\label{tab:cofw}
\end{table}

\minisection{Comparison with the state-of-the-art} We use the COFW set~\cite{burgos2013robust} for proving the effectiveness of our method. COFW consists of $500$ labeled images for training and $507$ for testing. Labels consist of binary masks. \cref{tab:cofw} reports our results compared to the state-of-the-art, along with ablation study to motivate our choices. The table reports figures for face IOU intersection over union (or Jaccard index), pixel accuracy (acc.), pixel-wise recall of the face class (rec$_\text{\emph{face}}$),\footnote{Starting from~\cite{ghiasi2015using}, only the rec$_\text{\emph{face}}$ has been reported on COFW omitting rec$_\text{\emph{all}}$; since a \emph{single} recall class can be made arbitrarily high by just optimizing the system for that class, we strove to report both for fairness.} average pixel-wise recall across \emph{all classes} (rec$_\text{\emph{all}})$ face and non-face, our measure of sparsity ($\frac{1}{N}\sum_i^N \vert\on{cc}_{\phi}-\on{cc}_{gt}\vert$) and fps (frames per second). When we test our method we simply merge the responses from the occlusion class and background class as a single non-face class. Following previous work~\cite{jia2014structured,Yang_tip_land,ghiasi2015using}, we report the metrics in the face box provided with COFW. Given the small size of COFW, it is challenging to fine-tune a deep model. To prove this point, and, more importantly, to motivate \cref{sec:tk}, we train from random weights (\emph{\small{+Scratch}}). Since we are updating the weights very slowly, the model is able to learn, yet reaches a result that is too distant from the state-of-the-art.
For this reason, previous methods~\cite{saito2016real,nirkin2018_faceswap} employed other labeled sets~\cite{GLOC_CVPR13} or built semiautomatic annotation tools~\cite{nirkin2018_faceswap} to attain some sort of transfer learning. Similar to them, we perform transfer learning, yet unlike them, we transfer knowledge from \cite{nirkin2018_faceswap} as explained in \cref{sec:tk}. Results in \cref{tab:cofw} (\emph{\small{+Transf.}}) support our face segmentation transfer. Our method is able to outperform the teacher network~\cite{nirkin2018_faceswap}. Additionally, if we combine all our novelties and further fine-tune on COFW, we obtain an additional positive gap with respect to the state-of-the-art (Struct. via con. \emph{\small{+Transf.+F.t.+reg.}}). Our method reduces the overall error-rate by \emph{27.7\%} for the metric rec$_\text{\emph{all}}$. As a final note, since we are using a lightweight encoder-decoder, unlike
\cite{saito2016real}, our smoothness constraint is enforced at \emph{training time only}. Our inference time is remarkable: on average a forward pass takes $3.1$ ms yielding more than $300$ predicted masks per second (fps).

\minisection{Ablation study} The effect of learning with ``structure via consensus'' is shown in \cref{tab:cofw} and is compared to the softmax+CE. While fine-tuning with the pixel-wise loss increases sparsity on the masks (1.0 $\rightarrow$ 3.8) and actually \emph{reduces performance}; on the contrary, by enforcing smoothness with our loss, we are able to better generalize to the test set, to improve over the transfer learning and to keep a lower sparsity (1.6). Further gain is obtained by regularizing the model with dropout and flip augmentation (\emph{+reg.}).
A qualitative comparison is shown in \cref{fig:cofw_qual}, where our method shows more structured masks than the baseline and \cite{nirkin2018_faceswap}. Other qualitative samples are shown in \cref{fig:teaser,fig:valid_loss:c,fig:qual_cofw_beta}.

\subsection{\partlabels~Database}\label{sec:partlabel}

\begin{table}[t]
\centering
\resizebox{.95\linewidth}{!}{
\begin{tabular}{l c@{~}@{~}c@{~}cc}
\toprule
{Method} & size$_{\emph{in}}$ & No CRF & acc.$_{\emph{p}}$ &  acc.$_{\emph{sp}}$  \\
\cmidrule(r){1-1} \cmidrule(l){2-5}
Gygli \etal \cite{gygli2017deep} --- DVN  & 32 &  \checkmark  & --- & 92.44  \\
Gygli \etal \cite{gygli2017deep} --- FCN baseline  & 32 & \checkmark   & --- & 95.36  \\
Kae \etal \cite{GLOC_CVPR13} --- CRF & 250 & \xmark & --- & 93.23 \\ %
Kae \etal \cite{GLOC_CVPR13} --- Glog & 250 & \xmark & --- & 94.95 \\ %
Liu \etal~\cite{liu2015multi} & 250 & \xmark & 95.24 & ---\\ %
Liu \etal~\cite{liu2017face} --- RNN & 128 & \checkmark & 95.46 & --- \\ %
Liu \etal~\cite{liu2017face,chen2014semantic} --- CNN-CRF & 128 & \xmark & 92.59 & --- \\
Saxena \etal (sparse)~\cite{saxena2016convolutional} &  250 & \checkmark  & 94.60 & 95.58  \\ %
Saxena \etal (dense)~\cite{saxena2016convolutional} & 250 &  \checkmark  & 94.82 & 95.63   \\ %
Zheng \etal~\cite{zheng2015learning} --- CNN-VAE & 250  & \checkmark & --- & 96.59\\ %
Tsogkas \etal~\cite{tsogkas2015deep} --- CNN & 250 & \checkmark & --- & 96.54\\ %
Tsogkas \etal~\cite{tsogkas2015deep} --- RBM+CRF & 250 & \xmark & --- &  \textbf{96.97}\\ %
Lin \etal~\cite{lin2019face} --- FCN+Mask-R-CNN  & 250 & \checkmark & \tbf{96.71} & --- \\
\ifx \Advtraining \undefined
\cmidrule(r){1-5} 
\else
\cmidrule(r){1-5}
\multicolumn{5}{c}{Adversarial Training}\\
FCN  --- GAN \cite{goodfellow2014generative} & 250 & \checkmark   & --- & 95.53  \\
GAN \cite{goodfellow2014generative} & 250 & \checkmark    & --- & 95.54  \\
FCN --- LSGAN \cite{Mao2016} & 250 & \checkmark    & --- & 95.51  \\
LSGAN \cite{Mao2016} & 250 & \checkmark    & --- & 95.52  \\
FCN --- WGAN,GP \cite{gulrajani2017improved} & 250 & \checkmark    & --- & 95.59  \\
WGAN,GP \cite{gulrajani2017improved} & 250 & \checkmark    & --- & 95.59  \\
FCN --- EBGAN \cite{zhao2016energy} & 250 & \checkmark    & --- & 95.50  \\
EBGAN \cite{zhao2016energy} & 250 & \checkmark     & --- & 95.52  \\
FCN --- LDRSP \cite{pan2018learning} & 250 & \checkmark    & --- & 95.87  \\
LDRSP \cite{pan2018learning} & 250 & \checkmark   & --- & {96.47} \\ %
\cmidrule(r){1-5}
\fi
Structure via Consenus (Ours) & 128 & \checkmark  &  \underline{96.05} & \underline{96.80}\\
Structure via Consenus (Ours) & 250 & \checkmark  &  {95.86} & 96.78\\
\bottomrule
\end{tabular}
}
\caption{{\bf \partlabels~set.} The comparison of pixel and super-pixel accuracies (acc.$_{\emph{p}}$, acc.$_{\emph{sp}}$). The input size and usage of smoothness via CRF are emphasized. Best result is in bold, while second best is underlined.
\vspace{-20pt}
}
\label{tab:lfw}
\end{table}

\minisection{Comparison with the state-of-the-art} Following previous work~\cite{GLOC_CVPR13}, we employ the funneled version of the set, in which images have already been coarsely aligned. Part Labels is a subset of LFW~\cite{LFWTech} for face segmentation proposed in~\cite{GLOC_CVPR13}, and consists of 1,500 training, 500 validation, and 927 testing images. The images are labeled with efficient super-pixel segmentation. The set provides three classes--background, hair/facial-hair and face/neck along with the corresponding super-pixel mapping. We fine-tune our system on the 2,000 train/val images and test on the 927 evaluation faces following the publicly available splits. When fine-tuning, we associate the \partlabels~classes with the same semantic class of the transfer learning except for the occlusion class being mapped to the new hair class.
To have a thorough comparison with current work, we report both pixel-wise (acc.$_{\emph{p}}$) and super-pixel-wise accuracies (acc.$_{\emph{sp}}$). To report the super-pixel accuracy, we select the most frequent predicted label in a super-pixel. Our system reports results on par with the state-of-the-art, noting that in our case we perform \emph{direct inference} (no CRF \checkmark), and we are \emph{not forcing any smoothness} via CRF at test-time.
\cref{tab:lfw} shows the state-of-the-art evaluation. We have results similar to Tsogkas \etal~\cite{tsogkas2015deep}, yet they use a CRF to smooth out the result. Notably, our approach shows similar numbers when compared with the active research of adversarial training (following the extensive experimentation from~\cite{pan2018learning}), though this latter requires more parameters because of the discriminator. \cref{tab:partlabel-fmeasure} reports also the F1-score following the recent work in~\cite{lin2019face}. Although our method works at a 128p resolution we report results at 250p by up-sampling the predictions with nearest neighbor interpolation. %

\begin{figure}[tbh]
    \centering
     \small{\hspace{-15pt}Input, Label Mask\hspace*{45pt}
     Baseline\hspace*{45pt}
     \tbf{Ours}\\}
    \includegraphics[width=.9\linewidth]{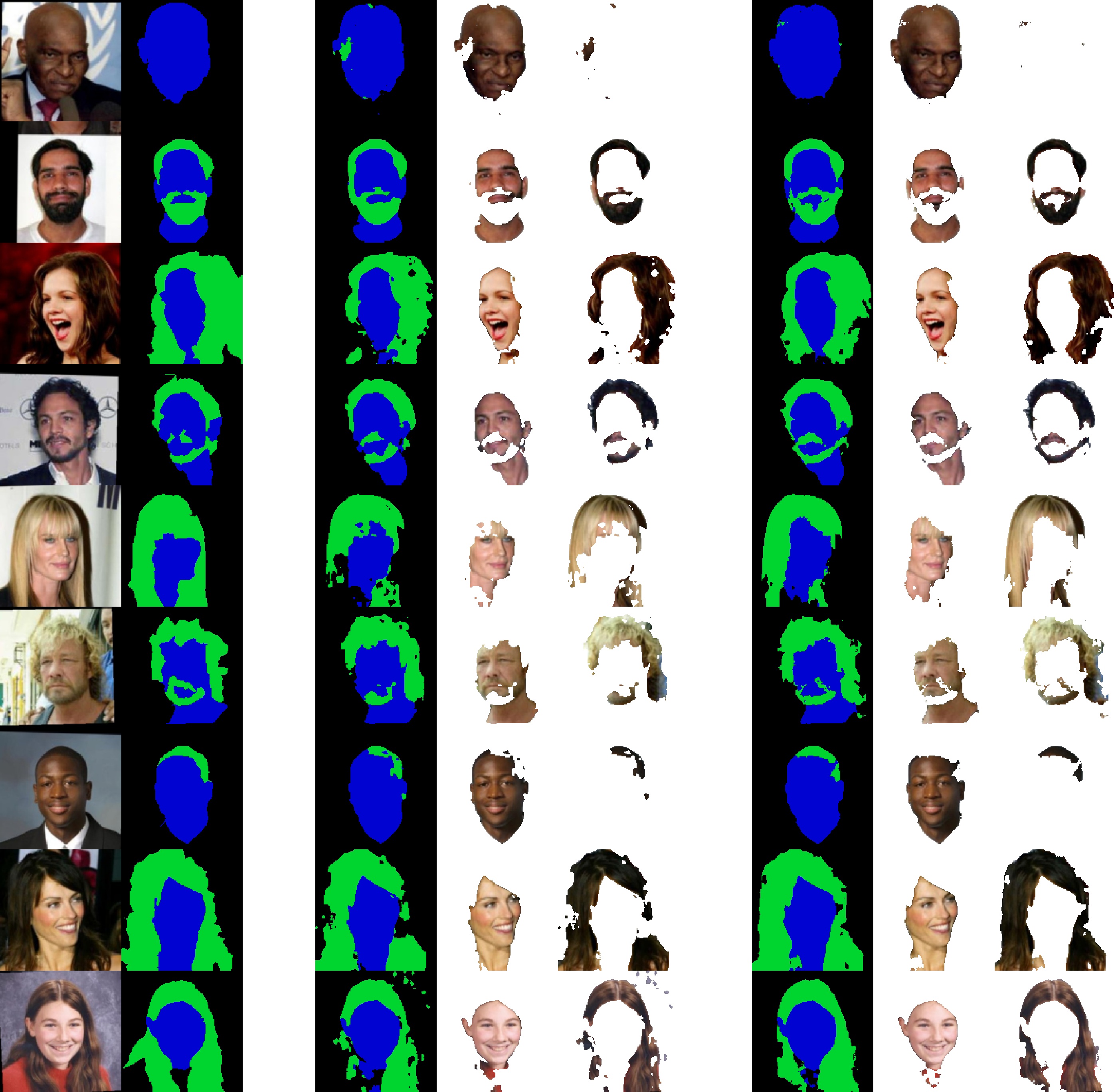}
    \caption{{\bf Qualitative results from the Part Labels.} Input image and its ground-truth mask; results by the baseline with pixel-wise loss; our result. The faces are masked to decouple the face from the hair. Additional results in the supplementary material in \cref{fig:partlabel_qual_supp}.
    \vspace{-25pt}}
    \label{fig:partlabel_qual}
\end{figure}

\minisection{Ablation Study} In \cref{tab:ablation-partlabel} we report ablation study showing the impact of our loss: in general pixel accuracy increases with our loss but since these metrics do not take into account class frequencies, we also recorded the IOU per class. Using ``structure via consensus'' the IOU for hair class goes up from $68.95\%$ to $72.48\%$. The same is reflected in the mean IOU over classes---from $83.65\%$ to $85.74\%$. We repeated the same experiments further regularizing the model with dropout and flip augmentation (\emph{\small{+reg.}}), our loss provided a similar improvement, and, importantly, the boost is consistent in all the metrics. Notably in all these ablations, our method provided less sparse masks when compared to the baseline as reported in \cref{tab:ablation-partlabel} under the sparsity metric, exhibiting less over-fitting than the baseline.
Qualitative results are shown in \cref{fig:partlabel_qual}: our hair segmentation exhibits less fragmented segments and fewer holes than the baseline, yet yielding an excellent face segmentation.

\begin{table}[t]
\centering
\resizebox{1.0\linewidth}{!}{
\begin{tabular}{l c@{~}c@{~}c@{~}c@{~}c@{~}c@{~~}c@{~}c@{~}}
\toprule
{Method} & IOU$_\text{\emph{hair}}$ &  IOU$_\text{\emph{bg}}$ & IOU$_\text{\emph{face}}$ & IOU$_\text{\emph{mean}}$ & recall$_\text{\emph{all}}$ & acc.$_{\emph{p}}$ &  acc.$_{\emph{sp}}$ & spars. \\
\cmidrule(r){1-1} \cmidrule(l){2-9}
Baseline                              & 68.95 & 94.41 & 87.60 & 83.65 & 90.41 & 94.77 & 96.15 & 15.86\\ %
Struct. via cons.                       & \tbf{72.48} &  \tbf{95.17} & \tbf{89.98} & \tbf{85.74} & \tbf{91.26} & \tbf{95.55} & \tbf{96.61} & 13.66\\ %
\cmidrule(r){1-1} \cmidrule(l){2-9}
Baseline \emph{\small{+reg.}}             & 73.97 & 95.52 & 89.81 & 86.46 & 92.50 & 95.77 & 96.62 & \tbf{3.3}\\ %
Struct. via cons. \emph{\small{+reg.}}  & \tbf{75.84} &  \tbf{95.74} & \tbf{90.62} & \tbf{87.40} & \tbf{93.22} & \tbf{96.05} & \tbf{96.80} & \tbf{3.3}\\ %
\bottomrule
\end{tabular}
}
\vspace{-5pt}
\caption{{\bf Ablation study on \partlabels~set.} Detailed ablation study on the \partlabels~set for the base model and a model with additional regularization (\emph{\small{+reg.}}).}
\vspace{-15pt}
\label{tab:ablation-partlabel}
\end{table}

\begin{table}[t]
\centering
\resizebox{.9\linewidth}{!}{
\begin{tabular}{l ccccc}
\toprule
{Method} & size$_{\emph{in}}$ & F1$_\text{\emph{face}}$ & F1$_\text{\emph{hair}}$ & F1$_\text{\emph{bg}}$ & acc.$_{\emph{p}}$  \\
\cmidrule(r){1-1} \cmidrule(l){2-6}
Liu \etal~\cite{liu2015multi}  & -- &    93.93 & 80.70 & 97.10 & 95.12 \\
Long \etal~\cite{long2015fully} & -- & 92.91 & 82.69 & 96.32 & 94.13 \\
Chen \etal~\cite{chen2018deeplab}  & --   & 92.54 & 80.14 & 95.65 &  93.44 \\
Chen \etal~\cite{chen2016semantic}  & --    & 91.17 & 78.85 & 94.95 & 92.49 \\
Zhou \etal~\cite{zhou2017face} & 320    & 94.10 & 85.16 & 96.46 & 95.28 \\
Liu \etal~\cite{liu2017face} & 128    & \tbf{97.55} & 83.43 & 94.37 & 95.46 \\
Lin \etal~\cite{lin2019face} & 250 & \underline{95.77} & \tbf{88.31} & \tbf{98.26} & \tbf{96.71} \\
\cmidrule(r){1-1} \cmidrule(l){2-6}
Struct. via cons. (Ours) & 128  & 95.08 & \underline{86.26} & \underline{97.82} & \underline{96.05}  \\
Struct. via cons. (Ours) & 250  & 94.74 & 85.74 & 97.72 & 95.86 \\
\bottomrule
\end{tabular}
}
\vspace{-5pt}
\caption{{\bf F1-score comparison on \partlabels.} Ours is the second best despite using a lightweight model. Numbers from~\cite{lin2019face}. Best result is in bold, while second best is underlined.}
\vspace{-8pt}
\label{tab:partlabel-fmeasure}
\end{table}

\section{Conclusions and Future Work}\label{sec:conclusions}
We proposed a novel method for face segmentation, building on the novel concept of learning structure via consensus. Our approach exhibits figures on par or above the state-of-the-art. Our future work is to experiment with Pascal VOC~\cite{everingham2010pascal} on the generic task of semantic segmentation, thereby porting our loss to work
with generic objects. The system is using blobs as a constraint for the consensus, and those are given as input to the system through an automatic, noisy preprocessing step or by some form of human supervision from the annotations. As a more long-term future work, we envision the possibility of learning to cluster
pixels of objects in an unsupervised fashion. 

\minisection{Acknowledgements} This research is based upon work supported by the Office of the Director of National Intelligence (ODNI), Intelligence Advanced Research Projects Activity (IARPA), via IARPA R\&D Contract No. 2017-17020200005. The views and conclusions contained herein are those of the authors and should not be interpreted as necessarily representing the official policies or endorsements, either expressed or implied, of the ODNI, IARPA, or the U.S. Government. The U.S. Government is authorized to reproduce and distribute reprints for Governmental purposes notwithstanding any copyright annotation thereon. The authors would like to thank S. Deutsch, A. Jaiswal, H. Mirzaalian, M. Hussein, L. Spinoulas and all the anonymous reviewers for their helpful discussions.
\appendix

\begin{figure*}[bth]
    \centering
    \includegraphics[width=.9\linewidth]{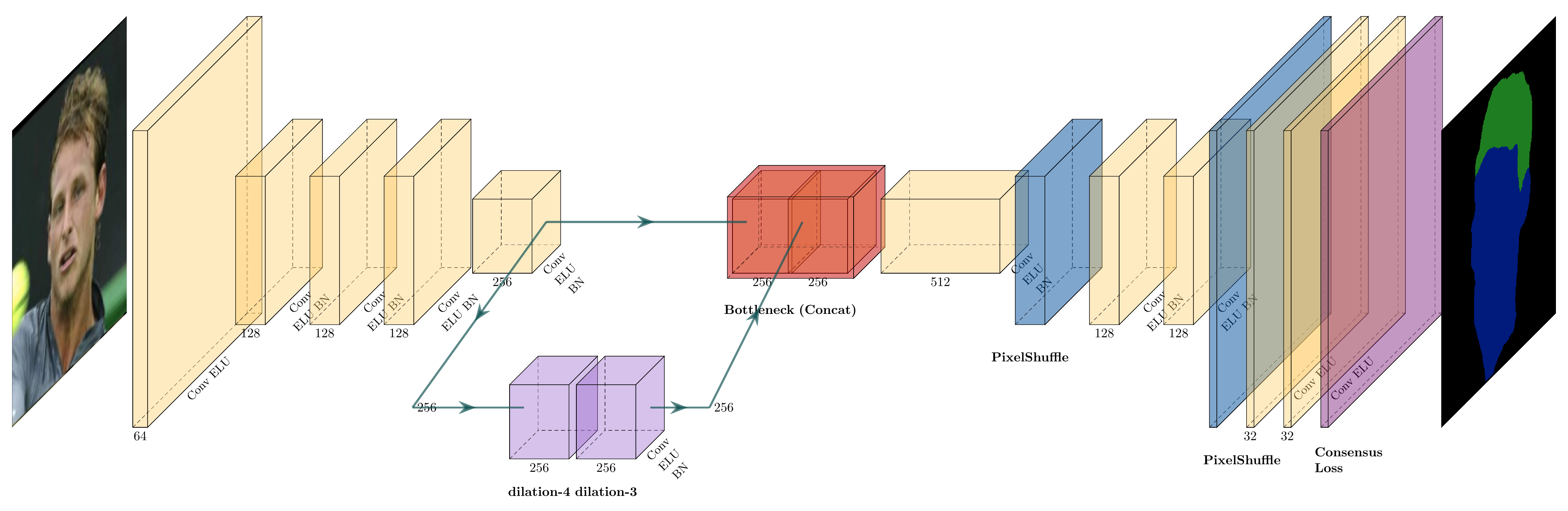}
    \caption{{\bf Network structure at a glance.} Encoder-decoder used for face segmentation and supervised through structure via consensus.}
    \label{fig:net}
\end{figure*}

\section{Network Architecture} \label{sec:supp_mat_net}
\minisection{Architecture Details} Further details of our network architecture are provided in \cref{tab:net}. Our network is similar to the encoder-decoder framework U-Net~\cite{ronneberger2015u}, but it has some modifications explained below. The input resolution is $128\PLH 128$, though our model can work also with other resolutions since it is fully convolutional. The resolution is decreased only with striding without any pooling layer. The basic block of the network consists of  \texttt{ReflectionPad2}, \texttt{Convolution}, \texttt{ELU}~\cite{clevert2015fast} and \texttt{Batch Normalization}. The first encoder downsamples the input up to $32\PLH32$ to preserve some spatial information, with a depth of 256 (layer id 19 referring to \cref{tab:net}); then other convolutional layers with dilation set to 4 and 3  in a sub-encoder refine the feature maps to capture a more global scale (layer id 27 referring to \cref{tab:net}). In \cref{tab:net}, if not specified, convolution is computed with dilation equal to 1. The final feature maps are concatenated together for a final bottleneck feature map with dimensionality $512\PLH32 \PLH 32$. The convolutional filters are initialized with the method described in ~\cite{glorot2010understanding}.
The decoder part takes the bottleneck feature maps as input and upscales it back to the input dimension. We used efficient sub-pixel convolution with ratio equal to 2 applied two times in the decoder to do this upscaling, since sub-pixel convolution has been shown to work well in super-resolution applications. We used Pytorch~\cite{paszke2017automatic} to develop the network and sub-pixel convolution has been implemented via \texttt{PixelShuffling}.\footnote{\href{https://pytorch.org/docs/stable/nn.html\#torch.nn.PixelShuffle}{pytorch.org/docs/stable/nn.html\#torch.nn.PixelShuffle}} The entire encoder-decoder has 4,524,323 parameters. The network is supervised either with 2D softmax normalization and cross-entropy or by using our novel ``structure via consensus'' method. The final network structure is displayed at a glance in \cref{fig:net} using~\cite{haris_iqbal_2018}.

\begin{table*}[p]%
\centering
\resizebox{.425\textheight}{!}{
    \begin{tabular}{l c@{~} @{\quad}c@{\quad}c@{~}}
    \toprule
ID    & Layer (type) &  Output Shape ($B\PLH C\PLH H\PLH W$) & Param. Size  \\
    \cmidrule(r){1-2} \cmidrule(l){3-4}
    \multicolumn{4}{c}{\tbf{Encoder} $\downarrow$}\\
1 &   ReflectionPad2d  &        [64, 3, 130, 130] &             ---\\  
2 &            Conv2d  &       [64, 64, 128, 128] &         1,792  \\ 
3 &               ELU  &       [64, 64, 128, 128] &             ---\\
\cmidrule(r){1-2}
4 &   ReflectionPad2d  &       [64, 64, 130, 130] &             ---\\
5 &            Conv2d  &        [64, 128, 64, 64] &        73,856  \\
6 &               ELU  &        [64, 128, 64, 64] &             ---\\
7 &       BatchNorm2d  &        [64, 128, 64, 64] &           256  \\
\cmidrule(r){1-2}       
8 &   ReflectionPad2d  &        [64, 128, 66, 66] &             ---\\
9 &            Conv2d  &        [64, 128, 64, 64] &       147,584  \\
10 &              ELU  &        [64, 128, 64, 64] &             ---\\
11 &      BatchNorm2d-  &        [64, 128, 64, 64] &           256  \\
\cmidrule(r){1-2}     
12 &  ReflectionPad2d-  &        [64, 128, 66, 66] &             ---\\
13 &           Conv2d  &        [64, 128, 64, 64] &       147,584  \\
14 &              ELU  &        [64, 128, 64, 64] &             ---\\
15 &      BatchNorm2d-  &        [64, 128, 64, 64] &           256  \\
\cmidrule(r){1-2}      
16 &  ReflectionPad2d  &        [64, 128, 66, 66] &             ---\\
17 &           Conv2d  &        [64, 256, 32, 32] &       295,168  \\
18 &              ELU  &        [64, 256, 32, 32] &             ---\\
19 &      BatchNorm2d  &        [64, 256, 32, 32] &           512  \\
\cmidrule(r){1-2}               
\multicolumn{4}{c}{\tbf{Sub-encoder} $\downarrow$}\\      
20 &  ReflectionPad2d  &        [64, 256, 40, 40] &             ---\\
21 &           Conv2d (dilation=4) &        [64, 256, 32, 32] &       590,080\\
22 &              ELU  &        [64, 256, 32, 32] &             ---\\
23 &      BatchNorm2d  &        [64, 256, 32, 32] &           512  \\
\cmidrule(r){1-2}       
24 &  ReflectionPad2d  &        [64, 256, 38, 38] &             ---\\
25 &           Conv2d (dilation=3)  &        [64, 256, 32, 32] &       590,080\\
26 &              ELU  &        [64, 256, 32, 32] &             ---\\
27 &      BatchNorm2d  &        [64, 256, 32, 32] &           512  \\
\cmidrule(r){1-2} 
\multicolumn{4}{c}{Concat feature maps 19 and 27} \\  
\cmidrule(r){1-2} 
\multicolumn{4}{c}{\tbf{Decoder} $\uparrow$} \\      
28 &  ReflectionPad2d  &        [64, 512, 34, 34] &             ---\\
29 &           Conv2d  &        [64, 512, 32, 32] &     2,359,808 \\
30 &              ELU  &        [64, 512, 32, 32] &             ---\\
31 &      BatchNorm2d  &        [64, 512, 32, 32] &         1,024  \\
32 &     \emph{PixelShuffle ($\PLH2$)}  &        [64, 128, 64, 64] &             ---\\
\cmidrule(r){1-2}         
33 &  ReflectionPad2d  &        [64, 128, 66, 66] &             ---\\
34 &           Conv2d  &        [64, 128, 64, 64] &       147,584  \\
35 &              ELU  &        [64, 128, 64, 64] &             ---\\
36 &      BatchNorm2d  &        [64, 128, 64, 64] &           256  \\
\cmidrule(r){1-2}         
37 &  ReflectionPad2d  &        [64, 128, 66, 66] &             ---\\
38 &           Conv2d  &        [64, 128, 64, 64] &       147,584  \\
39 &              ELU  &        [64, 128, 64, 64] &             ---\\
40 &      BatchNorm2d  &        [64, 128, 64, 64] &           256  \\
41 &     \emph{PixelShuffle ($\PLH2$)}  &       [64, 32, 128, 128] &             ---\\
\cmidrule(r){1-2}      
42 &  ReflectionPad2d &       [64, 32, 130, 130] &             ---\\
43 &           Conv2d &       [64, 32, 128, 128] &         9,248  \\
44 &              ELU &       [64, 32, 128, 128] &             ---\\
\cmidrule(r){1-2}               
45 &  ReflectionPad2d  &       [64, 32, 130, 130] &             ---\\
46 &           Conv2d  &       [64, 32, 128, 128] &         9,248  \\
47 &              ELU  &       [64, 32, 128, 128] &             ---\\
48 &           Conv2d  &        [64, 3, 128, 128] &           867  \\
           \cmidrule(r){1-1} \cmidrule(l){2-4}\\
     &      & Total \# params. & 4,524,323\\
    \bottomrule
    \end{tabular}
}
\caption{{\bf Network details.} Network layers, output shapes and learnable parameters.\vspace{2mm}}
\label{tab:net}
\end{table*}
\section{Additional Qualitative Results on COFW} \label{sec:supp_mat_cofw}
\begin{figure}[htb]
    \centering
    \small{\hspace{-13pt}Input, Label\hspace*{8pt}
     Nirkin \etal~\cite{nirkin2018_faceswap} \hspace*{5pt}
     Baseline \hspace*{15pt}
     \tbf{Ours}\hspace*{5pt}\\}
    \includegraphics[width=\linewidth,height=0.9\textheight,keepaspectratio]{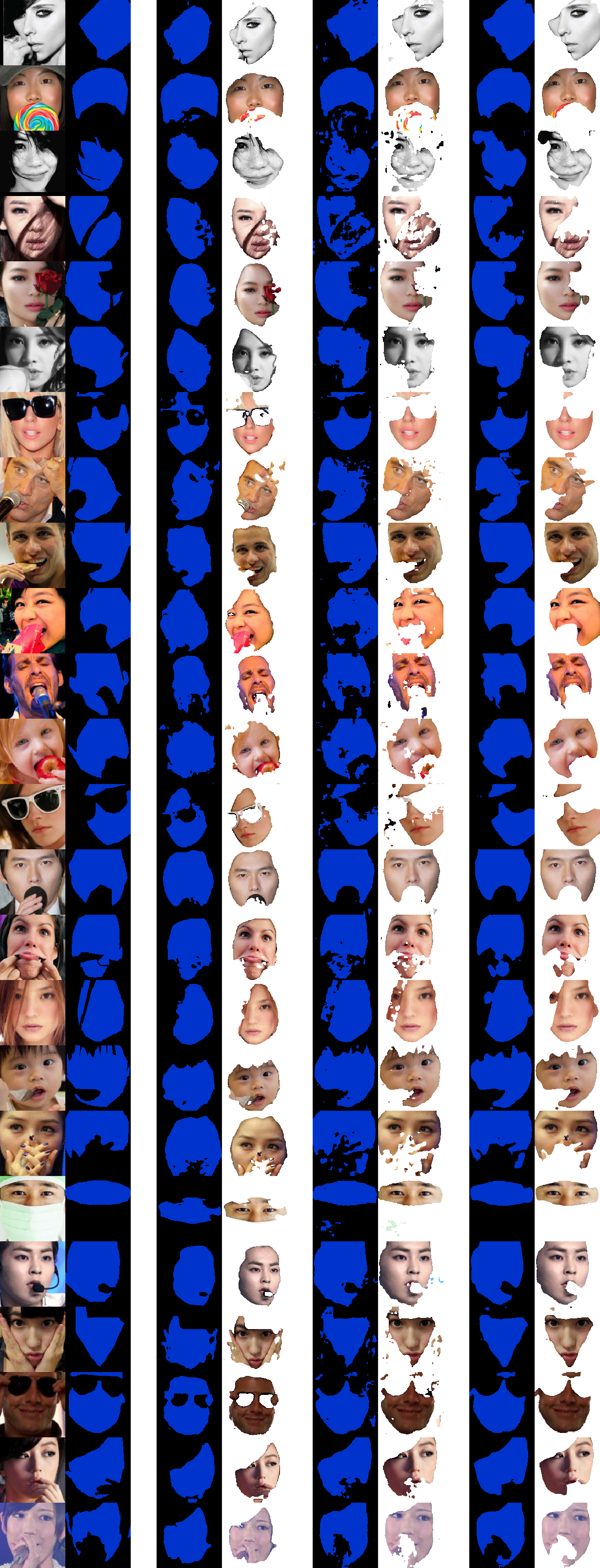}
    \caption{{\bf Additional qualitative samples from the COFW.} Input image and its ground-truth mask; results by Nirkin \etal~\cite{nirkin2018_faceswap}; baseline with pixel-wise loss; our result. The faces are masked to remove occlusions according to each method.}
    \label{fig:cofw1}
\end{figure}

We show supplementary results on the Caltech Occluded Face in the Wild data (COFW)~\cite{burgos2013robust} in \cref{fig:cofw1}.
The figures augment \cref{fig:cofw_qual} in the paper to provide further samples. The figures display the input image and its ground-truth mask; the result obtained by Nirkin \etal~\cite{nirkin2018_faceswap}, obtained by aligning the faces as the mentioned in their publicly available code; our baseline with pixel-wise softmax and cross-entropy; our final approach trained with structure via consensus.
\cref{fig:cofw1} show again that even on a larger pool of samples, our method returns less sparse, more continuous occlusion masks for better face segmentation and parsing. As a remark, we get such clean masks, much closer visually to the ground-truth compared to other approaches, yet we do so by \emph{still} performing pixel-wise inference: we do not use any super-pixel approach at test time nor employ any post-processing step such as CRF, morphological operations etc.

\section{Additional Qualitative Results\\ on Part Labels}\label{sec:supp_mat_pl}
\begin{figure}[tbh]
    \centering
     \small{\hspace{-35pt}Input, Label Mask\hspace*{17pt}
     Baseline\hspace*{25pt}
     \tbf{Ours}\\}
    \includegraphics[width=\linewidth,height=0.9\textheight,keepaspectratio]{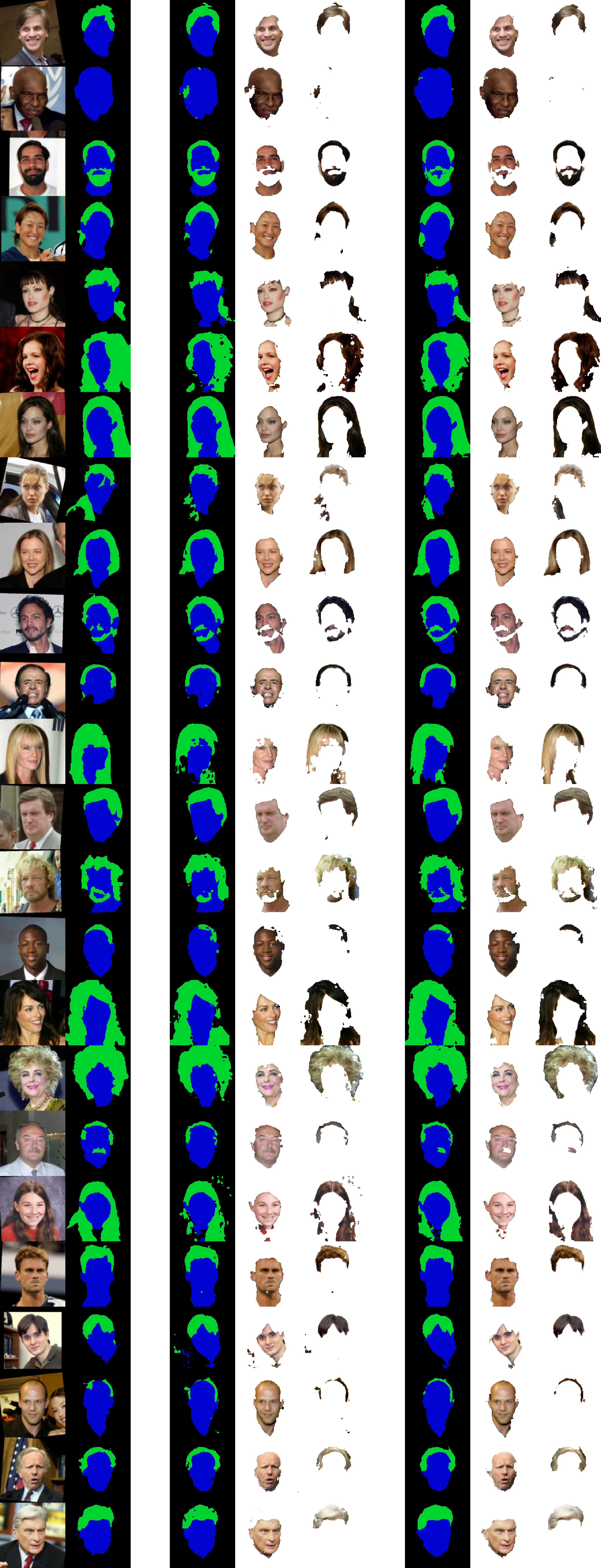}
    \caption{{\bf Additional qualitative samples from PartLabel.} Input image and its ground-truth mask; results by the baseline with pixel-wise loss; our result. The faces are masked to decouple the face from the hair.}
    \label{fig:partlabel_qual_supp}
\end{figure}

We show some supplementary qualitative results on the Part Labels database~\cite{GLOC_CVPR13} in \cref{fig:partlabel_qual_supp}. On average our masks look more continuous and greatly improve the IoU of the hair class. \cref{fig:partlabel_qual_supp} reports the input image, the ground-truth annotated mask, the baseline model trained with pixel-wise loss and regularization and our method with regularization. The result of each prediction for each class is used for segmenting part of the face showing the segmentation separately for face and hair.
In some cases, the predictions of our model are better than the super-pixel labels (e.g. tenth row).

\clearpage

\small{
\bibliographystyle{ieee_fullname}

}

\end{document}